\documentclass[letterpaper]{article} % DO NOT CHANGE THIS
\usepackage{aaai2026}  % DO NOT CHANGE THIS
\usepackage{times}  % DO NOT CHANGE THIS
\usepackage{helvet}  % DO NOT CHANGE THIS
\usepackage{courier}  % DO NOT CHANGE THIS
\usepackage[hyphens]{url}  % DO NOT CHANGE THIS
\usepackage{graphicx} % DO NOT CHANGE THIS
\usepackage{natbib}  % DO NOT CHANGE THIS AND DO NOT ADD ANY OPTIONS TO IT
\usepackage{caption} % DO NOT CHANGE THIS AND DO NOT ADD ANY OPTIONS TO IT
\usepackage{algorithm}
\usepackage{algorithmic}
\usepackage[most]{tcolorbox}
\usepackage{amsmath}
\usepackage{bm} 
\usepackage{amsfonts} 
\usepackage{booktabs}
\usepackage{multirow} 
\usepackage{xcolor}
\usepackage{colortbl}
\usepackage{paralist}

\definecolor{darkblue}{RGB}{44, 62, 80}     
\definecolor{mediumblue}{RGB}{52, 152, 219}   
\definecolor{lightblue}{RGB}{174, 214, 241}  
\definecolor{paleblue}{RGB}{235, 245, 251}    
\definecolor{grayblue}{RGB}{149, 165, 166}

\usepackage{newfloat}
\usepackage{listings}
\DeclareCaptionStyle{ruled}{labelfont=normalfont,labelsep=colon,strut=off} % DO NOT CHANGE THIS
\floatstyle{ruled}
\newfloat{listing}{tb}{lst}{}
\floatname{listing}{Listing}
\usepackage{adjustbox}
\title{\adjustbox{height=2.8em,valign=m}{\includegraphics{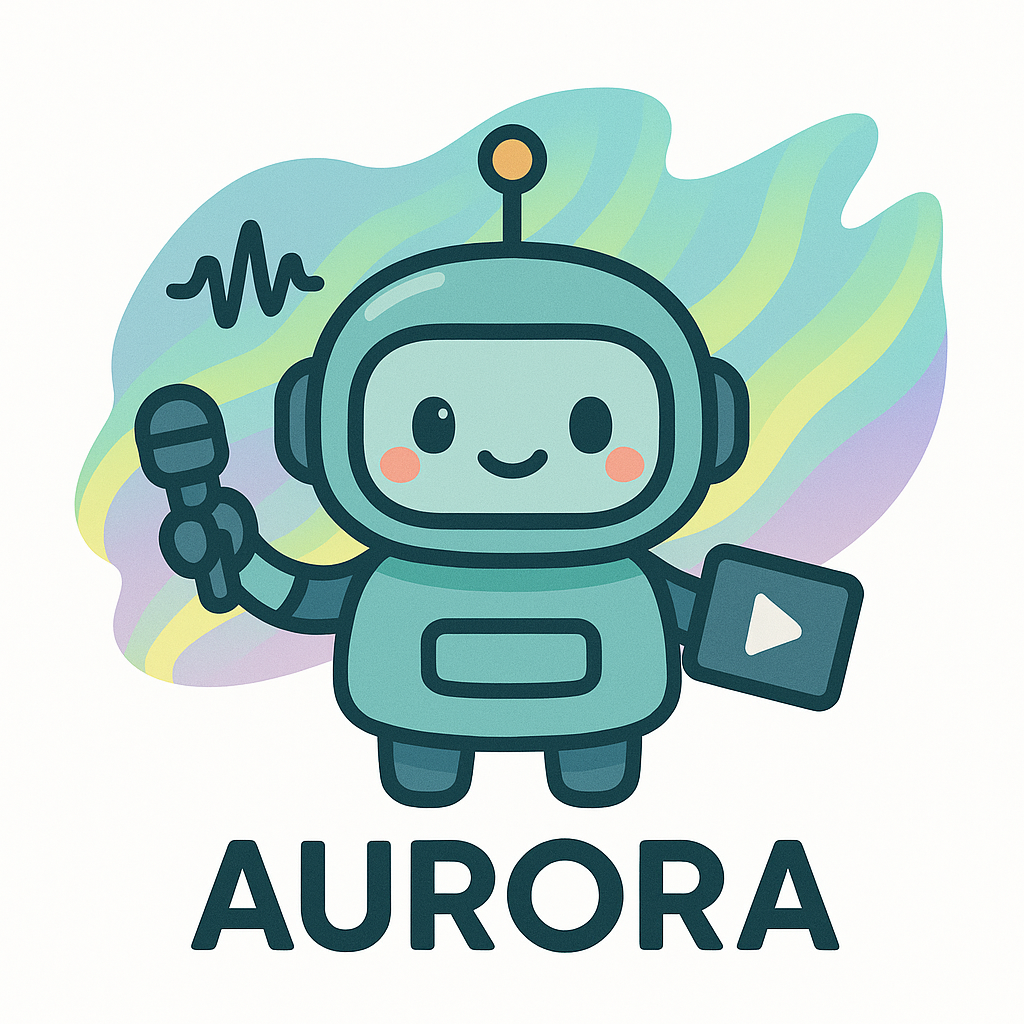}}\hspace{0.5em}AURORA: Augmented Understanding via Structured Reasoning and Reinforcement Learning for Reference Audio-Visual Segmentation}

\author {
    Ziyang Luo\textsuperscript{\rm 1}
    Nian Liu\textsuperscript{\rm 2}
    Fahad Shahbaz Khan\textsuperscript{\rm 2}
    Junwei Han\textsuperscript{\rm 1}
}
\affiliations {
    \textsuperscript{\rm 1}Northwestern Polytechnical University\\
    \textsuperscript{\rm 2}Mohamed bin Zayed University of Artificial Intelligence\\
}
\usepackage{bibentry}

\begin{document}

\maketitle

\begin{abstract}
Reference Audio-Visual Segmentation (Ref-AVS) tasks challenge models to precisely locate sounding objects by integrating visual, auditory, and textual cues. Existing methods often lack genuine semantic understanding, tending to memorize fixed reasoning patterns. Furthermore, jointly training for reasoning and segmentation can compromise pixel-level precision.
To address these issues, we introduce AURORA, a novel framework designed to enhance genuine reasoning and language comprehension in reference audio-visual segmentation. We employ a structured Chain-of-Thought (CoT) prompting mechanism to guide the model through a step-by-step reasoning process and introduce a novel segmentation feature distillation loss to effectively integrate these reasoning abilities without sacrificing segmentation performance. To further cultivate the model's genuine reasoning capabilities, we devise a further two-stage training strategy: first, a ``corrective reflective-style training" stage utilizes self-correction to enhance the quality of reasoning paths, followed by reinforcement learning via Group Reward Policy Optimization (GRPO) to bolster robustness in challenging scenarios. Experiments demonstrate that AURORA achieves state-of-the-art performance on Ref-AVS benchmarks and generalizes effectively to unreferenced segmentation.
\end{abstract}

\section{Introduction}

\begin{figure*}[!t]
    \centering
    \includegraphics[width=1\linewidth]{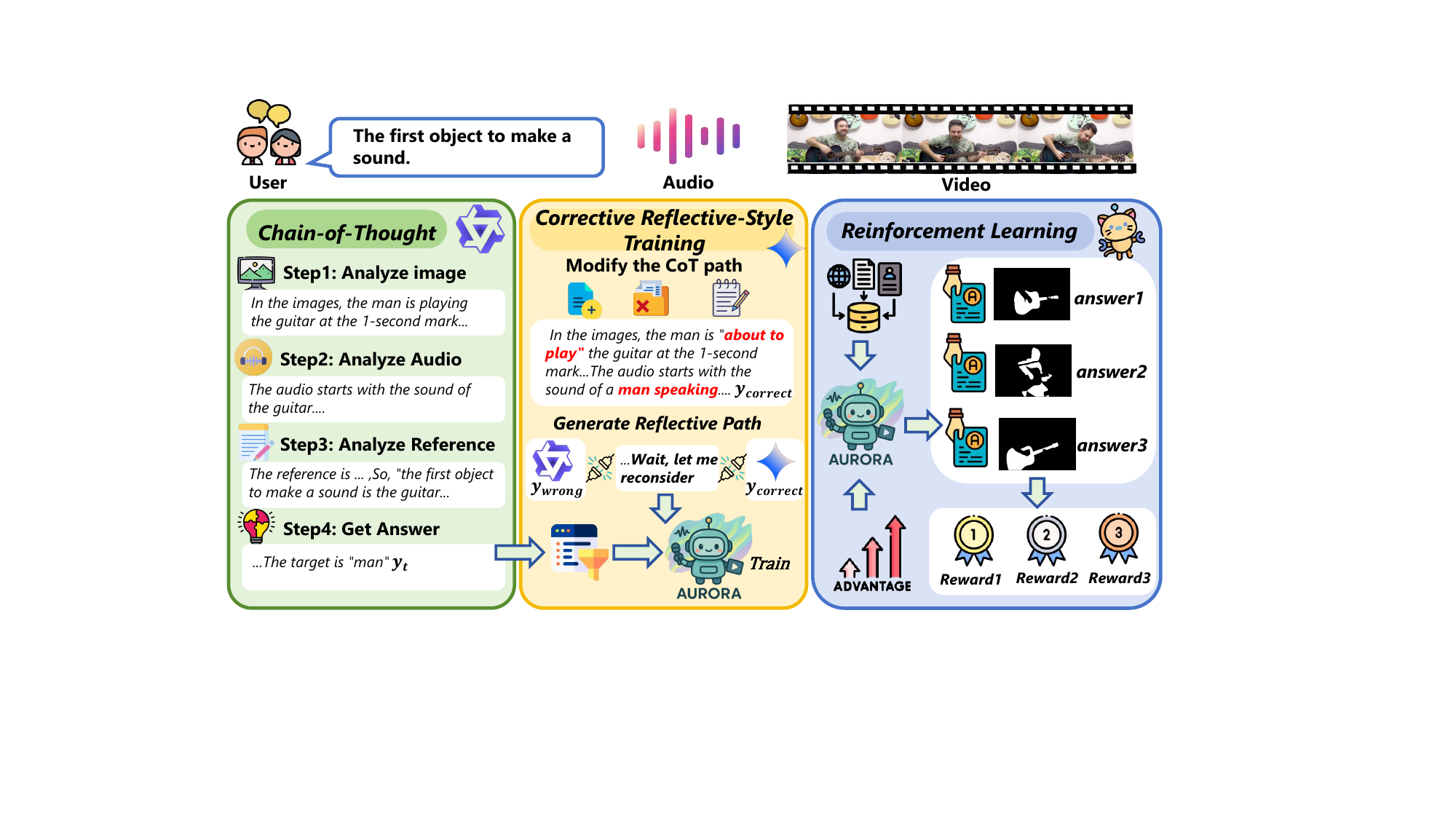}
    \caption{\textbf{Overall training pipeline of our proposed model.} The training pipeline consists of three stages: (1) \textbf{Supervised Fine-Tuning} with CoT paths ($\bm{y}_t$) generated by Qwen-Omni. (2) \textbf{Corrective Reflective-Style Training}, in which we construct a ``reflective-style'' path by combining the reasoning output from the SFT-trained model, a self-correction trigger, and the corrected path ($\bm{y}_{correct}$) from Gemini. (3) \textbf{Reinforcement Learning} via GRPO to further refine the model's reasoning.
    }
    \label{prompt_FIG}
    \vspace{-0.5cm}
\end{figure*} 

Humans perceive and interact with a complex, dynamic world by seamlessly integrating information from multiple sensory modalities. While vision plays a dominant role, auditory and textual cues are often crucial for accurately locating and comprehending objects of interest. This inherent multimodal nature of our perception highlights the necessity of developing systems that can achieve similar integrated understanding. Recent advancements in multimodal large language models, such as Qwen-Omni \cite{xu2025qwen2}, and VideoLLaMA2 \cite{cheng2024videollama}, have demonstrated significant progress in audio-visual comprehension. However, tasks like Reference Audio-Visual Segmentation (Ref-AVS) \cite{wang2024ref} still remain challenging, requiring models to precisely segment specific sounding objects by integrating textual references, audio cues, and visual information. This demands sophisticated multimodal reasoning alongside accurate spatial localization capabilities.

While previous works have explored fusion methods \cite{wang2024ref, wang2025sam2,radman2025tsam} to align modalities, often employing an auxiliary text encoder with self-attention fusion, it remains unclear whether these models truly grasp the meaning of textual references or merely exhibit language bias. The ``black box" nature of these approaches hinders genuine semantic understanding. 
To address these interpretability concerns, recent approaches have begun incorporating Large Language Models (LLMs) to enhance reasoning capabilities in Ref-AVS tasks, as demonstrated by Crab \cite{du2025crab}. 
However, their existing implementations often rely on supervised fine-tuning with pre-defined simple reasoning templates.
This approach tends to lead the LLM towards ``memorizing" these fixed patterns, which could result in outputs that reflect post-hoc rationalization rather than genuine and systematic reasoning. 
Moreover, the design of jointly training complex language reasoning with precise visual segmentation introduces new challenges, which risks compromising the pixel-level precision of the original segmentation models as pointed out by \cite{liu2025seg}.

In this paper, we propose AURORA,
a novel method that enhances both language comprehension and reasoning capabilities for audio-visual segmentation. Specifically, we leverage VideoLLaMA2 \cite{cheng2024videollama} as our MLLM component to work in tandem with SAM \cite{kirillov2023segment} for segmentation, enabling reasoning-guided segmentation. 
To prevent models from merely rationalizing the final output, we introduce a structured Chain-of-Thought (CoT) \cite{wei2022chain} prompting mechanism via open-source Qwen-Omni \cite{xu2025qwen2} to cost-effectively generate diverse reasoning paths for training. It involves distinct analytical steps focusing sequentially on visual, audio, and textual reference cues, followed by a final integration of these analyses to derive the answer. 
To mitigate the potential conflict between reasoning and segmentation during joint training, we introduce a segmentation feature distillation loss. This loss distills knowledge from a segmentation-only model into the joint model, enabling it to acquire CoT reasoning capabilities without compromising segmentation performance.
To endow the model with genuine reasoning ability, we further adopt a two-stage training strategy after supervised fine-tuning (SFT) with CoT, combining reflective learning and reinforcement learning via Group Reward Policy Optimization (GRPO) \cite{guo2025deepseek}. In the corrective reflective-style stage, we transform low-quality reasoning paths from SFT-trained model into ``reflective reasoning paths" by juxtaposing the initial flawed reasoning with a self-correction prompt followed by a Gemini-assisted high-quality revision. In the GRPO stage, we design a hybrid reward function incorporating format reward, IoU reward, and class reward to guide the model toward more robust and multimodally grounded reasoning.

Our main contributions can be summarized as follows:
\begin{compactitem}
\item We propose AURORA, a reasoning-enhanced Ref-AVS framework that introduces structured Chain-of-Thought prompting with segmentation feature distillation loss to enable genuine multimodal reasoning and preserve precise segmentation capabilities.
\item We further introduce a two-stage refinement approach: corrective reflective-style learning to mitigate foundation model biases in audio-visual understanding, followed by GRPO-based reinforcement learning for robust multimodal reasoning.
\item AURORA achieves state-of-the-art results on Ref-AVS benchmarks and generalizes effectively to unreferenced segmentation, demonstrating its robust reasoning and superior accuracy.
\end{compactitem}

\section{Related Work}
\subsection{Audio-Visual Segmentation}
Audio-Visual Segmentation (AVS) aims to generate pixel-level masks for sounding objects and has attracted increasing research attention \cite{zhou2023audio}. While some prior works have incorporated text modality into AVS, they primarily focus on class alignment without fully exploiting the semantic richness of textual information \cite{bhosale2023leveraging, bhosale2024unsupervised, luo2025tavis}. Furthermore, traditional AVS methods \cite{zhou2023audio,li2023catr,gao2024avsegformer,li2024qdformer,yang2024cooperation,ling2024transavs} lack explicit guidance mechanisms, making it challenging to identify specific objects of interest within complex audio-visual scenes.
To address this limitation, Ref-AVS \cite{wang2024ref} introduces a reference-based framework that provides continuous segmentation guidance through audio-visual-text fusion via a multi-modal cue aggregation module. Building upon this approach, SAM2-LOVE \cite{wang2025sam2} further enhances the framework by integrating SAM2 with multimodal fusion, token propagation, and accumulation strategies. TSAM \cite{radman2025tsam} enhances SAM with temporal modeling capabilities for spatio-temporal learning across video frames and replaces interactive prompting with data-driven prompts, thereby extending SAM's functionality to dynamic video content. However, these methods treat text as an auxiliary modality and may fail to fully understand reference semantics. 

Our work distinguishes itself by proposing a novel segmentation-centric framework integrating SAM \cite{kirillov2023segment} with VideoLLaMA2 \cite{cheng2024videollama} for reference-guided segmentation, employing CoT \cite{wei2022chain} reasoning to incrementally decompose references and a segmentation feature distillation loss to preserve segmentation accuracy in SFT training.

\subsection{Multimodal Language Models in Audio-Visual Scenes}
In audio-visual understanding, predominant Multimodal Language Models (MLLMs) such as Qwen-Omni \cite{xu2025qwen2}, and VideoLLaMA2 \cite{cheng2024videollama} integrate audio, visual, and textual modalities to achieve comprehensive scene understanding. MEERKAT \cite{chowdhury2024meerkat} advances this field by constructing a fine-grained, large-scale audio-visual instruction-tuning dataset that endows models with temporal and spatial grounding capabilities. More specifically for reference AVS, CRAB \cite{du2025crab} attempts to unify multimodal understanding and segmentation within an LLM framework. 
However, their reliance on supervised fine-tuning with pre-defined reasoning templates may lead to superficial pattern memorization rather than fostering genuine systematic reasoning or deep semantic comprehension, limiting their robustness in complex scenarios. 
To address these limitations, following SFT, we introduce an additional two-stage training process. The first stage, corrective reflective-style training, encourages the model to refine its prior knowledge and calibrate its foundational perceptual abilities. The second stage, GRPO, facilitates robust self-improvement by optimizing the model's reasoning pathways based on targeted reward functions.

\section{Methodology}
In this work, we introduce AURORA, a novel reasoning-enhanced framework for reference audio-visual segmentation that addresses the fundamental challenge of achieving genuine multimodal reasoning and maintaining precise segmentation performance.  
We implement this framework by integrating SAM \cite{kirillov2023segment} with VideoLLaMA2 \cite{cheng2024videollama}. 
Our approach employs a three-stage training procedure. In the first stage, we incorporate CoT reasoning using Qwen-Omni \cite{xu2025qwen2} for SFT with segmentation feature distillation loss. In the second stage, we enhance the model's foundational accuracy through a novel error-correction stage using structured ``reflective examples" that address common perceptual and knowledge errors. The third stage employs GRPO \cite{guo2025deepseek} to jointly enhance both reasoning capabilities and segmentation performance. 
Figure~\ref{prompt_FIG} illustrates the overall architecture of our proposed framework.

\vspace{-2mm}
\subsection{Stage1: SFT Training with CoT}
\label{stage1}
\subsubsection{CoT}
To fully utilize GRPO, the initial procedure involves teaching the model proper reasoning techniques and establishing the basic format. Without this foundation, the model can hardly perform effective self-improvement, leading to suboptimal optimization directions. Therefore, in the first training stage, we implement a cold start using SFT. 
However, simple reasoning like basic scene description or superficial audio analysis in isolation is insufficient,
as the model may engage in post-hoc rationalization of the final results, which could introduce complications in the subsequent GRPO procedure. To address this issue, we introduce CoT reasoning, which decomposes the reasoning process into step-by-step components rather than post-hoc explanations. 

Specifically, we first generate CoT answers using the open-source Qwen-Omni \cite{xu2025qwen2} model to conserve computational costs compared to proprietary models, and establish the initial format through carefully designed prompts. Through prompt engineering, we decompose the reference reasoning into four distinct steps as shown below.
\vspace{-2mm}
\begin{tcolorbox}[
    enhanced,
    colback=gray!8,
    colframe=black!60,
    colbacktitle=black!80,
    coltitle=white,
    title={\textsc{Chain of Thought Reasoning for Ref-AVS}},
    fonttitle=\bfseries\small,
    boxrule=1pt,
    titlerule=0pt,
    toptitle=4pt,
    bottomtitle=4pt,
    left=6pt,
    right=6pt,
    top=4pt,
    bottom=4pt
]
\renewcommand{\arraystretch}{1.2}
\small
\vspace{-1mm}
% \begin{tabular}{@{}p{0.3\textwidth}p{0.65\textwidth}@{}}
\textbf{Step 1 Video Description:} \textit{Extract important visual information from the video, including key actions, objects, and their spatial relationships.} \\
\textbf{Step 2 Audio Description:} \textit{Extract and analyze audio content, identifying sound classes, and temporal characteristics.} \\
\textbf{Step 3 Reference Analysis:} \textit{Analyze the relationship between multimodal information and the given reference, identifying relevant connections and correspondences.} \\
\textbf{Step 4 Final Answer:} \textit{Generate the final segmentation decision as ``the target is \{class\_name\}'' based on comprehensive analysis.}
% \end{tabular}
\end{tcolorbox}
The SFT objective aims to maximize the likelihood of a target CoT reasoning sequence $y=(y_1,...,y_T)$ conditioned on the input query $q$, which consists of the instruction  $\bm{x}_{inst}$, reference $\bm{x}_{r}$, video $\bm{v}$, and audio $\bm{a}$. The loss is defined as: 
\begin{equation} \label{sft}
\mathcal{L}_{SFT} = -\mathbb{E}_{(q, y) \sim \mathcal{D}} \sum_{t=1}^{T} \log \pi_\theta(y_t | q, y_{<t}),
\end{equation}
where $D$ denotes the fine-tuning dataset, and $\pi_\theta$ is the MLLM parameterized by $\theta$. Since Ref-AVS is a binary segmentation task, we follow the LISA framework \cite{lai2024lisa} and append the token ``\verb|It is [SEG]|'' after the CoT reasoning. The hidden state feature of the \verb|[SEG]| token is then used as a visual prompt for the SAM decoder.

\begin{figure*}[!t]
    \centering
    \includegraphics[width=1\linewidth]{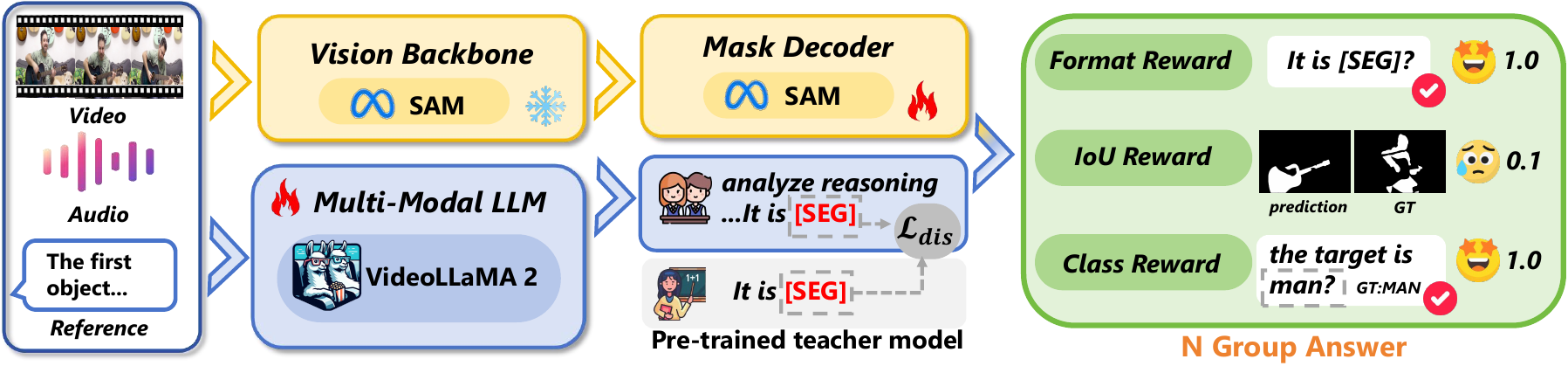}
   \caption{\textbf{Overall framework of our proposed model.} Our model integrates SAM and VideoLLaMA2. The gray block represents the Segmentation Feature Distillation Loss during the SFT stage, and the green block denotes the GRPO process with triplet rewards in Stage 3.}
    \label{prompt+seg}
    \vspace{-0.5cm}
\end{figure*}

\subsubsection{Segmentation Feature Distillation Loss}
While CoT reasoning has demonstrated effectiveness in language tasks, it presents a challenge for multi-modal segmentation tasks. We hypothesize that fine-tuning with a joint objective may compromise the pixel precision of the segmentation models \cite{liu2025seg}, particularly when the language-based reasoning loss dominates the optimization dynamics. This, in turn, can degrade the representation of the \verb|[SEG]| token, which is critical for triggering high-quality mask generation, thereby impairing segmentation performance. To address this issue and effectively decouple the optimization of reasoning and segmentation capabilities, we introduce a segmentation feature distillation loss as shown in the gray block in Figure~\ref{prompt+seg}. 

To obtain an ideal feature representation for distillation, we train a specialist teacher model. Unlike our main models, the teacher's training is deliberately confined to a pure segmentation task using only simple prompts (\verb|It is [SEG]|). This specialized training regime, involving more extensive training steps, allows it to develop a highly-optimized and uncompromised \verb|[SEG]| feature embedding ($f_t$), even though it entirely lacks complex reasoning capabilities. 
Subsequently, we train a student model with the CoT objective, while compelling its  \verb|[SEG]| feature embedding $f_s$, to mimic the teacher's representation by minimizing the Mean Squared Error (MSE) between them: 
\begin{equation} \label{distill}
\mathcal{L}_{dis} = \text{MSE}(f_t, f_s).
\end{equation}
Please note that we use the SAM decoder from the pre-trained teacher model and freeze its parameters during the student's training.
This strategy prevents potentially conflicting gradients from the complex reasoning task from altering the core segmentation module, ensuring that the student model acquires CoT reasoning skills without compromising the segmentation fidelity inherited from the teacher.

\subsection{Stage2: Corrective Reflective-Style Training}
\label{stage2}
While SFT with CoT data enables models to generate reasoned outputs and offers a cost-effective starting point, we identify a critical limitation: models often struggle with foundational perceptual and knowledge-based errors, such as misinterpreting audio cues (e.g., confusing the volume of a violin and a cello; \textit{\textbf{detailed analysis is provided in the Supplementary Material}}). Simply fine-tuning on correct CoT examples may not efficiently rectify these deep-seated biases.
To address this, we introduce a corrective reflective-style training scheme. These structured examples explicitly present a common error followed by a detailed correction, effectively creating a ``reflective'' learning signal.
Rather than teaching abstract self-reflection, this method's primary function is to refine the model's prior knowledge and calibrate its foundational perceptual abilities. 

To endow our model with a more robust reasoning capability, we require high-quality examples of ideal thinking processes. We leverage a more powerful MLLM, Gemini, to serve as an expert annotator, whose role is to generate gold-standard reasoning paths by correcting flawed CoT samples.
Specifically, we first prompt Gemini to critique the reasoning generated by our Stage 1 model. This process is focused on challenging samples from the training set, which we identify as those with both an IoU score below 0.6 and incorrect reasoning.
This strict criterion ensures that corrective training targets only clear-cut failures where both the reasoning process and segmentation outcome are flawed.
Adhering to the minimal modification principle \cite{yu2024rlhf}, Gemini is instructed to rectify any identified flaws—such as incorrect conclusions or superficial analysis of audio information—by making the edits (e.g., modifying, adding, or deleting words). We denote the initial, potentially flawed CoT from the Stage 1 model as $y_{wrong}$ and the corrected version from Gemini as $y_{correct}$. We then construct a ``reflective path" as follows:
\begin{equation} \label{eq:reflective_path}
y_{reflective} = y_{wrong} \oplus  x_{trigger}\oplus y_{correct},
\end{equation}
where $x_{trigger}$ is a reflective trigger phrase randomly sampled from a predefined collection (e.g., ``Wait, let me re-evaluate..."). This path explicitly models the cognitive process of identifying a flaw and subsequently correcting it. Finally, we perform a second stage of fine-tuning on the model weights from our SFT stage. In this stage, we replace the standard target sequences $y_{t}$ with our newly constructed $y_{reflective}$ paths, thereby directly injecting the desired ``reflective'' behavior into the model.

\begin{table*}[t!]
    \setlength{\belowcaptionskip}{0.2cm}   
    \renewcommand{\arraystretch}{1.3}     
    \renewcommand{\tabcolsep}{9pt}        
    \small                                 
    \centering
	\begin{tabular}{l@{\hspace{1.5em}}ccc@{\hspace{2em}}ccc@{\hspace{2em}}ccc}
		\toprule
		\multirow{2}{*}{\textbf{Method}} & \multicolumn{3}{c}{\textbf{Seen}}  & \multicolumn{3}{c}{\textbf{Unseen}} & \multicolumn{3}{c}{\textbf{Mix (S+U)}}\\
		\cmidrule(lr){2-4} \cmidrule(lr){5-7} \cmidrule(lr){8-10} 
		& \textbf{$\mathcal{J \uparrow}$} & \textbf{$\mathcal{F\uparrow}$} & \textbf{$\mathcal{J}\!\&\!\mathcal{F}\uparrow$} & \textbf{$\mathcal{J\uparrow}$} & \textbf{$\mathcal{F\uparrow}$} & \textbf{$\mathcal{J}\!\&\!\mathcal{F\uparrow}$} & \textbf{$\mathcal{J\uparrow}$} & \textbf{$\mathcal{F}\uparrow$} & \textbf{$\mathcal{J}\!\&\!\mathcal{F}\uparrow$} \\
		\midrule
		\rowcolor{paleblue} \multicolumn{10}{l}{\textit{Audio-based Segmentation Methods}} \\
		\addlinespace[0.1em]
        \quad AVSBench \cite{zhou2022audio}  & 23.2 & 51.1 & 37.2 & 32.4 & 54.7 & 43.5 & 27.8 & 52.9 & 40.3 \\
        \quad AVSegFormer \cite{gao2024avsegformer} & 33.5 & 47.0 & 40.2 & 36.1 & 50.1 & 43.1 & 34.8 & 48.6 & 41.7\\
        \quad GAVS \cite{wang2024prompting} & 28.9 & 49.8 & 39.4 & 29.8 & 49.7 & 39.8 & 29.4 & 49.8 & 39.6\\ 
        \addlinespace[0.3em]
         \rowcolor{paleblue} \multicolumn{10}{l}{\textit{Visual-based Segmentation Methods}} \\
        \addlinespace[0.1em]
        \quad ReferFormer \cite{wu2022language} & 31.3 & 50.1 & 40.7 & 30.4 & 48.8 & 39.6 & 30.9 & 49.5 & 40.2\\
        \quad R2VOS \cite{li2023robust} & 25.0 & 41.0 & 33.0 & 27.9 & 49.8 & 38.9 & 26.5 & 45.4 & 35.9\\
        \addlinespace[0.3em]
         \rowcolor{paleblue} \multicolumn{10}{l}{\textit{Multi-modal Methods}} \\
        \addlinespace[0.1em]
        \quad EEMC \cite{wang2024ref} & 34.2 & 51.3 & 42.8 & 49.5 & 64.8 & 57.2 & 41.9 & 58.1 & 50.0\\
        \quad SAM-LAVS \cite{wang2025sam2} & 43.5 & 51.9 & 47.7 & 66.5 & 72.3 & 69.4 & 55.0 & 62.1 & 58.5\\
        \quad TSAM \cite{radman2025tsam} &43.4 &56.8 & 50.1 &54.6 &66.4 &60.5 &49.0 &61.6  &55.3\\
        \rowcolor{paleblue} \multicolumn{10}{l}{\textit{Foundation-based Methods}} \\
        \addlinespace[0.1em]
        \quad CRAB \cite{du2025crab} &40.5 &58.0 &49.3 &45.6 &63.0 &54.3 &43.1 &60.5 &46.2\\
        % \midrule
          \rowcolor{paleblue} \textbf{AURORA(Ours)} &\textbf{63.2} &\textbf{72.8} &\textbf{68.0}&\textbf{69.7} &\textbf{76.4} &\textbf{73.0} &\textbf{66.5} &\textbf{74.6} &\textbf{70.1}\\
        
		\bottomrule
	\end{tabular}
    \caption{\textbf{Performance comparison across different methods in Seen, Unseen, and Mix (S+U) settings of Ref-AVS benchmark. } The mix
indicates the average value of seen and unseen splits. ``$\uparrow$'' indicates higher is better. }
   \label{SOTA}
\end{table*}

\subsection{Stage3: Reinforcement Learning Training}
\label{stage3}
Following the SFT stage, we incorporate reinforcement learning (RL) to further enhance the reasoning capability of the MLLM. Specifically, we adopt the GRPO \cite{guo2025deepseek} framework to enable self-refinement based on the relative quality of generated outputs within a group. However, GRPO was originally designed for text generation and does not directly align with the segmentation nature of Ref-AVS. To address this, we redesign the reward function tailored for segmentation as shown in the green block of Figure~\ref{prompt+seg}.

\subsubsection{GRPO}
Given an input query $q$, GRPO samples a group of  $G$ candidate responses $o=\{o_1,...o_G\}$ from the current policy $\pi_\theta$. A rule-based reward model evaluates each response to produce scalar rewards $\{R_1,...,R_G\}$. To assess the relative quality within the group, each reward is standardized to compute a normalized advantage for the $i$-th response:
\begin{equation}\label{advantage}
\hat{A}_i = \frac{R_i - \text{mean}(\{R_i\}_{i=1}^G)}{\text{std}(\{R_i\}_{i=1}^G)}.
\end{equation}
The optimization objective balances improving response quality with maintaining proximity to the previous policy via a clipped objective.
The final GRPO loss is defined as: 
\begin{align}
\mathcal{L}_{\text{GRPO}}(\theta) &= \mathbb{E}_{(q,a)\sim\mathcal{D},\{o_i\}_{i=1}^G\sim\pi_{\theta_{\text{old}}}(\cdot|q)} \left[ \frac{1}{G} \sum_{i=1}^{G} \frac{1}{|o_i|} \sum_{t=1}^{|o_i|} \right. \notag \\
&\quad \left. \min \left( r_{i,t}(\theta)\hat{A}_{i,t}, \text{clip} \left( r_{i,t}(\theta), 1 - \varepsilon, 1 + \varepsilon \right) \hat{A}_{i,t} \right) \right. \notag \\
&\quad \left. - \beta D_{\text{KL}}(\pi_\theta || \pi_{\text{ref}}) \right],
\end{align}
where  ${r}_{i,t} (\theta)= \frac{\pi_{\theta}(o_{i,t}|q,o_{i,<t})}{\pi_\text{old}(o_{i,t}|q,o_{i,<t})}$. Following \cite{yu2025dapo}, we set the KL divergence coefficient $\beta$ to zero, effectively removing this term for simplicity.
\subsubsection{Reward Design}
To effectively guide AURORA, we design a hybrid reward function consisting of Format Reward $R_{format}$, IoU Reward $R_{IoU}$, and Class Reward $R_{class}$. 
The coefficients for these reward components are all set to 1.

\paragraph{\textbullet\ Format Reward}
The format reward evaluates whether the output reasoning adheres to the required structural format. In tasks like mathematical problem-solving, reasoning is often followed by a variable-content textual answer enclosed in specific placeholders, such as \verb|<answer></answer>|. While for our segmentation task, the goal is different. Although the model is still expected to generate a textual reasoning chain, the process must end with a fixed trigger phrase that directly leads to the generation of the segmentation mask. Therefore, we implement a simpler format constraint: if the final sentence of the reasoning process is \verb|"It is [SEG]"|, the format reward is assigned a value of 1; otherwise, it receives a value of 0. This approach not only aligns conceptually with the segmentation task but also avoids introducing extraneous tokens absent from the vocabulary of our base model. 

\paragraph{\textbullet\ IoU Reward}
Text-based rewards primarily optimize for linguistic coherence and semantic correctness, but they cannot directly assess the accuracy of the generated segmentation mask. Therefore, we introduce an IoU-based reward to measure segmentation mask quality. Specifically, we calculate the IoU between the predicted segmentation mask and the ground truth mask, which ranges from 0 to 1, to generate the reward signal, addressing a critical limitation that cannot be resolved through text generation rewards alone. 

\paragraph{\textbullet\ Class Reward}
Simply introducing the IoU reward cannot directly supervise the correctness of reasoning; therefore, we introduce a class reward to evaluate the reasoning procedure. Since our CoT reasoning includes step 4, which outputs \verb|"the target is {class_name}"|,
we extract the \verb|{class_name}| and compare it with the true class. If the \verb|{class_name}| is consistent with the true class, we set the reward to 1; otherwise, we set it to 0. For simplicity, we do not consider synonyms in this evaluation.

\section{Experiment}
\subsection{Experimental Settings}
\subsubsection{Dataset}
We evaluated our method on the Ref-AVS benchmark dataset \cite{wang2024ref}, which contains 4,000 videos with manual pixel-level annotations and expressions. 
The dataset is divided into a training set (2,908 videos), a validation set (276 videos), and a test set (818 videos). The test set is further split into three subsets: (1) seen split: containing the same 39 categories as the training set; (2) unseen split: with 13 additional categories not present during training; (3) null split: testing cases where expressions refer to non-existent or invisible objects. As in \cite{wang2025sam2}, we omit the null subset from evaluation.
\vspace{-2mm}
\subsubsection{Implementation Details}
Our training unfolds in three stages. First, we perform an initial SFT for 720 steps using CoT reasoning generated by Qwen-Omni \cite{xu2025qwen2}. In the second stage, we conduct a 300-step corrective fine-tuning phase. For this, we curate 1,505 reflective-style examples by using Gemini 2.5-Pro to refine incorrect outputs from the first stage. These are then mixed with 3,500 standard CoT samples to prevent catastrophic forgetting. The final stage consists of 500 steps of GRPO \cite{guo2025deepseek} training, where we generate 3 candidate responses per input for preference alignment. We use a batch size of 2 with 1 sample per device and gradient accumulation steps of 4. All experiments are conducted on two A40 GPUs.

\vspace{-2mm}
\subsubsection{\textbf{Evaluation Metrics}}
In line with the evaluation protocol of Ref-AVS \cite{wang2024ref}, we employ the Jaccard index ($\mathcal{J}$), F-score ($\mathcal{F}$), and their average ($\mathcal{J} \& \mathcal{F}$) as the primary evaluation metrics.

\subsection{Main Results}
We conduct a comprehensive performance comparison of our proposed AURORA against a suite of top-performing SOTA methods on the Ref-AVS benchmark. These include three audio-based segmentation methods \cite{zhou2022audio, gao2024avsegformer, wang2024prompting}, two visual-based segmentation methods \cite{wu2022language, li2023robust}, three multi-modal methods \cite{wang2024ref, wang2025sam2, radman2025tsam}, and one foundation-based method \cite{du2025crab}. The detailed results, presented in Table~\ref{SOTA}, demonstrate that AURORA achieves new state-of-the-art performance across multiple key metrics.

\begin{figure*}[!t]
    \centering
    \includegraphics[width=1\linewidth]{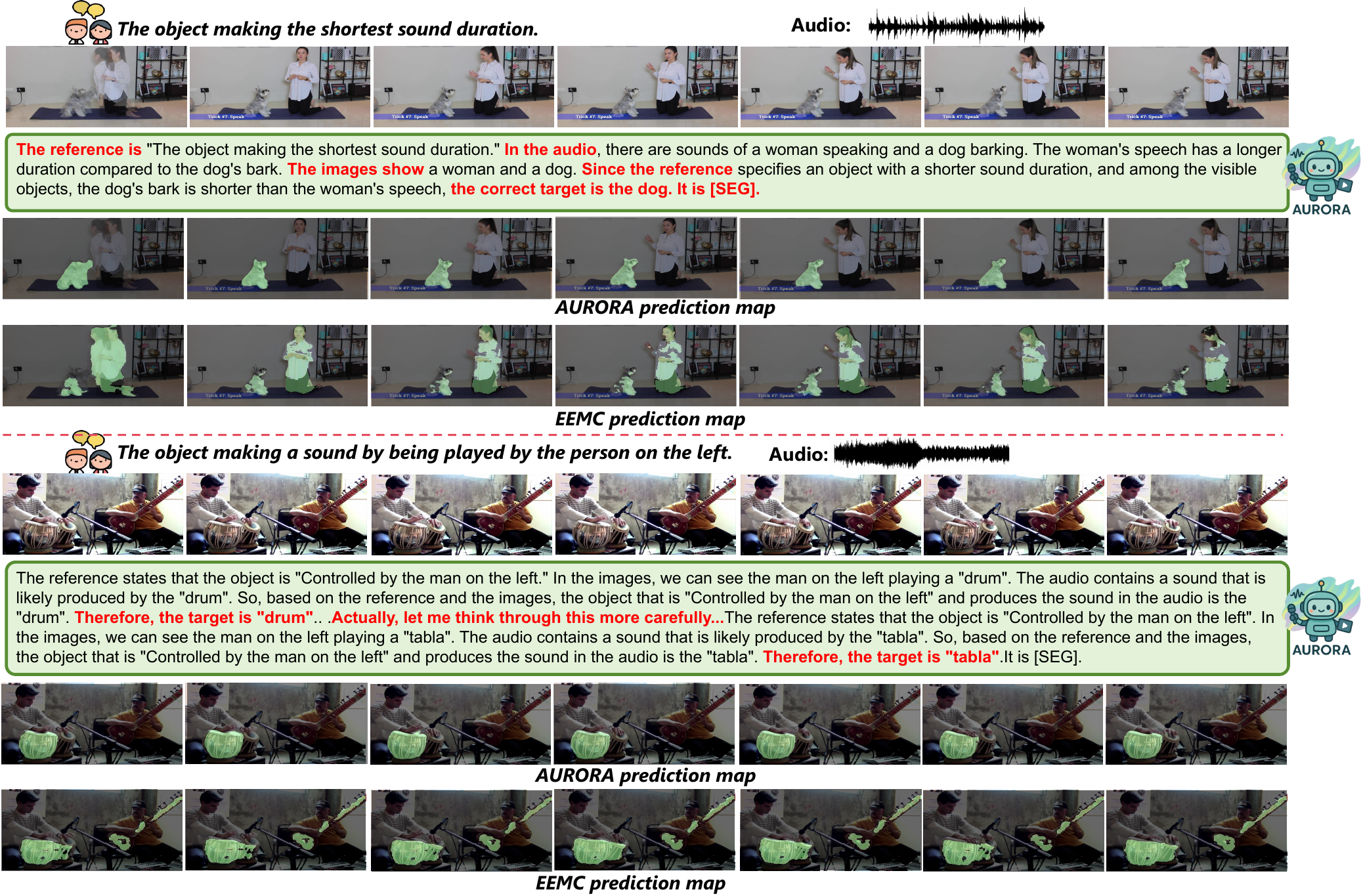}
    \vspace{-0.7cm}
    \caption{\textbf{The visualization results of the referred objects in the Ref-AVS compared with EEMC \cite{wang2024ref}.} Note that although the reasoning steps may appear in different orders to enhance diversity and support GRPO exploration, all outputs consistently contain the \textbf{four key reasoning steps} shown in Figure~\ref{prompt_FIG}. }
    \label{compare}
    \vspace{-0.5cm}
\end{figure*}

On the seen test split, our model surpasses the second-best performing method by substantial margins. We attribute this significant leap in performance to our framework's core design: unlike prior methods that rely on simple modality fusion \cite{wang2024ref, wang2025sam2, radman2025tsam} or train segmentation models from scratch \cite{du2025crab}, AURORA leverages the power of a large language model for deep semantic reasoning and integrates it with a pre-trained foundation model for segmentation. More importantly, AURORA demonstrates even greater advantages on the challenging unseen test split. This consistent improvement on unseen categories indicates that AURORA possesses enhanced generalization capabilities, enabling it to effectively locate and segment novel objects not present in the training data.
Figure~\ref{compare} provides visual comparisons among the top-performing models.

\subsection{Ablation Study}

\begin{table}[t!]
    \setlength{\belowcaptionskip}{0.2cm}   
    \renewcommand{\arraystretch}{1}     
    \renewcommand{\tabcolsep}{6pt}         
    \small                                   
    \centering
	\begin{tabular}{l@{\hspace{1.5em}}ccc@{\hspace{1em}}ccc}
		\toprule
		\multirow{2}{*}{\textbf{Method}} & \multicolumn{3}{c}{\textbf{Seen}}  & \multicolumn{3}{c}{\textbf{Unseen}} \\
		\cmidrule(lr){2-4} \cmidrule(lr){5-7} 
		& \textbf{$\mathcal{J}$} & \textbf{$\mathcal{F}$} & \textbf{$\mathcal{J}\!\&\!\mathcal{F}$} & \textbf{$\mathcal{J}$} & \textbf{$\mathcal{F}$} & \textbf{$\mathcal{J}\!\&\!\mathcal{F}$} \\
        \midrule
        w/o MLLM &39.2 &48.7 &44.0 &34.0  &46.3 &40.2\\
        w/o CoT &54.1 &64.0 &59.1 &64.2 &71.9 &68.1\\
        \rowcolor{paleblue} CoT &\textbf{61.4} &\textbf{70.6} &\textbf{66.0} &\textbf{67.1} &\textbf{73.4} &\textbf{70.3}\\
		\bottomrule
	\end{tabular}
    \vspace{-3mm}
    \caption{\textbf{Ablation study of SFT training with CoT.}} 
    \label{SFT}
    \vspace{-4mm}
\end{table}

\begin{table}[t!]
    \setlength{\belowcaptionskip}{0.2cm}   
    \renewcommand{\arraystretch}{1}      
    \renewcommand{\tabcolsep}{7pt}         
    \small                                
    \centering
	\begin{tabular}{l@{\hspace{1.5em}}ccc@{\hspace{1em}}ccc}
		\toprule
		\multirow{2}{*}{\textbf{Method}} & \multicolumn{3}{c}{\textbf{Seen}}  & \multicolumn{3}{c}{\textbf{Unseen}} \\
		\cmidrule(lr){2-4} \cmidrule(lr){5-7} 
		& \textbf{$\mathcal{J}$} & \textbf{$\mathcal{F}$} & \textbf{$\mathcal{J}\!\&\!\mathcal{F}$} & \textbf{$\mathcal{J}$} & \textbf{$\mathcal{F}$} & \textbf{$\mathcal{J}\!\&\!\mathcal{F}$} \\
        \midrule
        w/o $\mathcal{L}_{dis}$ &{60.2} &{70.2} &{65.2} &{65.7} &{72.4} &{69.1}\\
        \rowcolor{paleblue} with $\mathcal{L}_{dis}$ &\textbf{61.4} &\textbf{70.6} &\textbf{66.0} &\textbf{67.1} &\textbf{73.4} &\textbf{70.3} \\
		\bottomrule
	\end{tabular}
    \vspace{-3mm}
    \caption{\textbf{Ablation study of segmentation feature distillation loss.}} 
    \label{distill}
    \vspace{-6mm}
\end{table}

\begin{table}[t!]
    \setlength{\belowcaptionskip}{0.2cm}   
    \renewcommand{\arraystretch}{1}      
    \renewcommand{\tabcolsep}{4.5pt}         
    \small                                 
    \centering
	\begin{tabular}{l@{\hspace{1.5em}}ccc@{\hspace{1em}}ccc}
		\toprule
		\multirow{2}{*}{\textbf{Method}} & \multicolumn{3}{c}{\textbf{Seen}}  & \multicolumn{3}{c}{\textbf{Unseen}} \\
		\cmidrule(lr){2-4} \cmidrule(lr){5-7} 
		& \textbf{$\mathcal{J}$} & \textbf{$\mathcal{F}$} & \textbf{$\mathcal{J}\!\&\!\mathcal{F}$} & \textbf{$\mathcal{J}$} & \textbf{$\mathcal{F}$} & \textbf{$\mathcal{J}\!\&\!\mathcal{F}$} \\
        \midrule
        SFT+GRPO &62.0 & 71.5 &66.8 &69.1 &76.1 &72.6\\
        SFT+CT+GRPO &62.7 &72.3 &67.5  &67.7 &74.4 &71.1\\
        \rowcolor{paleblue}SFT+CRT+GRPO &\textbf{63.2} &\textbf{72.8} &\textbf{68.0} &\textbf{69.7} &\textbf{76.4} &\textbf{73.0} \\
		\bottomrule
	\end{tabular}
    \vspace{-3mm}
    \caption{\textbf{Ablation study of corrective reflective-style training}. 
    Here, CT denotes fine-tuning on corrected paths ($y_{correct}$), and CRT is our reflective training on full paths ($y_{reflective}$).
    } 
    \label{reflective}
    \vspace{-5mm}
\end{table}

\subsubsection{Ablation study of SFT training with CoT}
To evaluate our method, we first compare it against a non-MLLM baseline where an ImageBind \cite{girdhar2023imagebind} text encoder processes simple prompts for SAM \cite{wang2025sam2}. As shown in Table~\ref{SFT} (w/o MLLM), our full model achieves superior performance, demonstrating the benefits of using a powerful MLLM backbone.
To validate the contribution of CoT, we conduct an ablation study by training a variant of our model without the CoT component (w/o CoT). The resulting performance drop confirms that CoT is crucial for effective multimodal reasoning.

\begin{table}[t!]
    \setlength{\belowcaptionskip}{0.2cm}   
    \renewcommand{\arraystretch}{1}     
    \renewcommand{\tabcolsep}{2.4pt}         
    \small                              
    \centering
	\begin{tabular}{l@{\hspace{1em}}ccc@{\hspace{1em}}ccc}
		\toprule
		\multirow{2}{*}{\textbf{Method}} & \multicolumn{3}{c}{\textbf{Seen}}  & \multicolumn{3}{c}{\textbf{Unseen}} \\
		\cmidrule(lr){2-4} \cmidrule(lr){5-7} 
		& \textbf{$\mathcal{J}$} & \textbf{$\mathcal{F}$} & \textbf{$\mathcal{J}\!\&\!\mathcal{F}$} & \textbf{$\mathcal{J}$} & \textbf{$\mathcal{F}$} & \textbf{$\mathcal{J}\!\&\!\mathcal{F}$} \\
        \midrule
        with Reflective  &{61.8} &{71.1} &{66.5} &{67.2} &{73.9} &{70.6} \\
       $R_{format}$  + $R_{IoU}$ &62.4 &72.2 &67.3 &68.3 &75.4 &71.9\\
        \rowcolor{paleblue} $R_{format}$  + $R_{IoU}$ +$R_{class}$ &\textbf{63.2} &\textbf{72.8} &\textbf{68.0} &\textbf{69.7} &\textbf{76.4} &\textbf{73.0}\\
		\bottomrule
	\end{tabular}
    \vspace{-3mm}
    \caption{\textbf{Ablation study of different designs in GRPO stage.} } 
    \label{reward}
    \vspace{-5mm}
\end{table}

\subsubsection{Effectiveness of Segmentation Feature Distillation Loss}
While our joint training approach is effective, it can create tension between the segmentation and reasoning tasks as they compete for shared parameters. 
Thus, we employ a feature distillation loss to instill robust segmentation capabilities by transferring knowledge from a specialist segmentation model.
The results in Table~\ref{distill} confirm that this loss maintains segmentation performance without harming reasoning, as evidenced by the results on the unseen split.

\subsubsection{Effectiveness of Corrective Reflective-Style Training}
We evaluate our corrective reflective-style training by its final performance after our two-stage pipeline, as shown in Table~\ref{reflective}. Our pipeline begins with an SFT stage to learn the $y_{reflective}$ format, followed by a GRPO stage to optimize this capability (SFT+CRT+GRPO).
First, to demonstrate the value of incorporating reflective data, we compare our full pipeline against a baseline trained without any reflective examples (SFT+GRPO). The substantial performance gains of our method validate the overall effectiveness of our approach.
Moreover, we test if performing SFT with only the correct answer $y_{correct}$ is enough (SFT+CT+GRPO). Our full reflective-style training still performs better. We believe this is because the reflective path explicitly shows the model both the mistake and the correct answer. This is more effective for rectifying the model's inherent biases than simply training on the correct answer alone.

\subsubsection{Ablation study of GRPO}
We evaluate the effectiveness of our GRPO stage in Table~\ref{reward}, using the Stage 2 reflective model as our baseline (with Reflective). Our GRPO training first incorporates an $R_{IoU}$ reward, which yields substantial improvements over the reflective model. To further enhance performance on difficult samples, we introduce an additional $R_{class}$ reward, which achieves further gains. These progressive improvements confirm that our reward design effectively enhances segmentation by refining the model's reasoning abilities. For all GRPO experiments, a simple $R_{format}$ reward was used as a training aid to ensure the model consistently generates the required \verb|[SEG]| token, but it does not directly contribute to the final performance metrics.

\begin{table}[t!]
    \setlength{\belowcaptionskip}{0.2cm}   
    \renewcommand{\arraystretch}{1}      
    \renewcommand{\tabcolsep}{25.9pt}         
    \small                             
    \centering
	\begin{tabular}{l@{\hspace{3em}}cc}
		\toprule
		\multirow{2}{*}{\textbf{Method}} & \multicolumn{2}{c}{\textbf{S4}}   \\
		\cmidrule(lr){2-3} 
		& \textbf{$\mathcal{J}$} & \textbf{$\mathcal{F}$} \\
        \midrule
        Crab  &73.3 &86.8 \\
        \rowcolor{paleblue} \textbf{AURORA} &77.3 &86.7\\
		\bottomrule
	\end{tabular}
    \vspace{-3mm}
    \caption{\textbf{Comparison with the fine-tuned AVS method \cite{du2025crab} on the S4 subset of AVSBench.}} 
    \label{zero-shot}
    \vspace{-5mm}
\end{table}

\subsection{Cross-Task Generalization Analysis}
To evaluate the transferability of our model's learned representations, we assessed its performance on the AVSBench dataset \cite{zhou2022audio}, which focuses on generic audio-visual segmentation without reference guidance. This task represents a significant departure from our training paradigm, as it requires identifying salient sounding objects rather than following specific textual references.
To adapt our model, we guided the model with the generic prompt, ``The sounding object," and exclusively fine-tuned our GRPO-trained model for 3 epochs. Crucially, we did not generate new CoT samples or reflective paths for the fine-tuning process. 
As shown in Table~\ref{zero-shot}, our adapted model achieves competitive performance on the $\mathcal{J}$ metric compared to the fully fine-tuned Crab model \cite{du2025crab}, which also builds upon a MLLM with reasoning capability.
Importantly, our model autonomously applies its learned reasoning capabilities to the new task through GRPO-evolved reasoning patterns, demonstrating genuine transferable reasoning rather than task-specific memorization.
Note that we omit MS3 evaluation due to our use of a single \verb|[SEG]| token, suited for single-object segmentation.

\section{Conclusion} 
We propose AURORA, a framework that endows Ref-AVS with authentic reasoning. 
First, we introduce a structured CoT prompting mechanism during SFT to build a strong foundation for reasoning.
To mitigate the conflict between reasoning and segmentation during joint training, we introduce a feature distillation loss that preserves pixel-level precision.
To elevate the model's capabilities from simple rationalization to authentic introspection, we further develop a two-stage refinement strategy combining reflective learning and GRPO-based reinforcement learning.  AURORA achieves state-of-the-art performance on Ref-AVS benchmarks and generalizes well to the AVS task.

\bibliography{main}

\clearpage

\section{More Details of Proposed Method}
\subsection{Qwen-Omni Prompt for Chain-of-thought}
As introduced in Stage 1 of the main paper, we leverage the capabilities of the advanced MLLM, Qwen-Omni \cite{xu2025qwen2}, to generate the initial dataset with CoT reasoning. To ensure the generated reasoning is structured, logical, and aligned with our task requirements, we designed a specific prompt to guide the model. This prompt not only provides the necessary multimodal context and task-specific information, but also establishes a clear, step-by-step format for the reasoning process. The complete Qwen-Omni prompt format used for this data generation process is detailed below. It instructs the model to produce a coherent explanation that logically connects the provided inputs to the final answer, using a provided example as a template for the desired output style.

\begin{tcolorbox}[
enhanced,
colback=gray!8,
colframe=black!60,
boxrule=1pt,
left=6pt,
right=6pt,
top=4pt,
bottom=4pt
]
\textit{I will provide you with an audio clip and three images sampled from a 10-second video, taken at 1s, 5s, and 10s, along with a reference statement. Additionally, I will give you the correct target answer. Your task is to generate a reasoning process that logically deduces the given target based on the provided audio, images, and reference. \\
\textbf{Reference}: ``xxxxxx" \\
\textbf{Target}: ``xxxxx" \\
\textbf{Example}:\\
\textbf{Reference}: ``The object that produces a sound of shorter duration than the handpan."\\
\textbf{Target}: ``Violin."\\
\textbf{Reasoning}: \\
The images show a person playing both the handpan and a violin.  The audio contains the sound of a handpan, which has a long, resonant duration. Additionally, another sound can be heard that is significantly shorter in duration. Since the reference specifies an object with a shorter sound duration compared to the handpan, and among the visible objects, the violin produces shorter-duration sounds, the correct target is "violin". \\
\textbf{Now, generate the reasoning process for the given Reference and Correct.}}
\end{tcolorbox}

\subsection{Gemini Prompt for Corrective Reflective-style Training}
As detailed in Stage 2 of the main paper, we employ a corrective reflective-style training scheme to address the deep-seated perceptual and knowledge-based errors observed in the initial SFT model. The core of this stage is the generation of high-quality ``reflective paths", which requires an expert annotator to correct the flawed reasoning produced by our Stage 1 model. To this end, we leverage the superior capabilities of Gemini as our expert annotator. The prompt requires Gemini to output both the original flawed reasoning with errors marked, and the corrected version with modifications highlighted. This structured output is essential for constructing the final reflective path, which explicitly teaches our model to recognize and amend its own mistakes. The complete Gemini prompt format is presented below.

\textit{You are an assistant designed to help me correct an incorrect answer to a question about a video with audio. I will provide you with 10 images which are sampled from a video, an audio, a reference, an answer from an LVLM, and the final correct result.
First, verify the reasoning and correct it if it's wrong. If the answer is right, just return ``right" as the final output. If the answer is wrong, you should modify, add, or delete certain words or sentences in the LVLM's answer to correct mistakes, including incorrect objects and faulty reasoning. \\
The corrected answer should be the final correct result.\\
You should return: \\
(1) The original LVLM's answer I provided, in which you should mark the deleted or modified parts in quotes.  
(2) Your corrected answer, in which you should mark the modified or added parts compared to the original answer in quotes.
Please ensure that only the modified, deleted, or added parts are marked. Do not mark synonyms as modifications. Please retain the sentence structure and content of the LVLM's original answer as much as possible, without adding extra information beyond what is necessary for correction. \\
I will give you an example:\\
(1) The original LVLM's answer:The reference is ``The first object to make a sound." In the images, the man is playing the guitar at the 1-second mark, and the dog is lying on the couch. ``The audio starts with the sound of the guitar." So, ``the first object to make a sound is the guitar, not the man." The target is ``man" but the correct answer ``should be ``guitar" based on the reference." \\
(2) Your corrected answer: The reference is ``The first object to make a sound." In the images, the man is ``about to play" the guitar at the 1-second mark, and the dog is lying on the couch. ``The audio starts with the sound of a man speaking, saying 'This is enjoy'." So, the first object to make a sound is the ``man". The target is ``man" and the correct answer is ``man" based on the reference.\\
Please note that for audio-related reasoning, you should focus on analyzing the sounds instead of directly giving meaningless reasons based on the correct result. For example, you should not just say The audio is of a musical performance. The piano's sound gets lower in volume over time. So, based on the reference and the images, the target is the ``piano".\\
You should describe the audio and analyze to get this answer. Now, give me the answer
}
\subsection{Overall Training Loss}
In Stage 1, we perform supervised fine-tuning using a CoT format and introduce four loss functions. The cross-entropy loss $\mathcal{L}_{ce}$ enables the model to learn CoT reasoning patterns. To preserve segmentation capabilities, we employ binary cross-entropy loss $\mathcal{L}_{bce}$ and Dice loss $\mathcal{L}_{dice}$ for pixel-level classification. Additionally, we incorporate a segmentation feature distillation loss $\mathcal{L}_{dis}$. This ensures that as the student model learns CoT reasoning via $\mathcal{L}_{ce}$, it simultaneously preserves the high-quality segmentation fidelity inherited from the teacher model. The overall loss function is formulated as:
\begin{equation} \label{total}
\mathcal{L}_{stage1} = \mathcal{L}_{{ce}} + \mathcal{L}_{{bce}} + \mathcal{L}_{{dice}} +  \mathcal{L}_{dis}.
\end{equation}
In Stage 2, we replace the standard CoT reasoning with our proposed ``reflective path" approach to address deep-seated knowledge biases and perceptual errors and train the model on these corrected paths using the cross-entropy loss. The training procedure follows a similar structure to Stage 1, with the exception that the distillation loss $\mathcal{L}_{dis}$ is removed since the model has already acquired the necessary segmentation knowledge from the teacher. The loss function is defined as:
\begin{equation} \label{total}
\mathcal{L}_{stage2} = \mathcal{L}_{{ce}} + \mathcal{L}_{{bce}} + \mathcal{L}_{{dice}}.
\end{equation}
In Stage 3, we introduce GRPO to enhance the model's reasoning capabilities. Unlike previous stages, we do not employ cross-entropy loss, as it may constrain the model's reasoning flexibility. Instead, we focus on preference-based optimization while maintaining segmentation performance. The loss function combines the GRPO loss with segmentation-specific losses:
\begin{equation} \label{total}
\mathcal{L}_{stage3} = \mathcal{L}_{{GRPO}} + \mathcal{L}_{{bce}} + \mathcal{L}_{{dice}}.
\end{equation}

\section{More Analysis}
\subsection{Analysis of Reflective dataset}
To empirically validate the central thesis presented in the main text—that SFT with CoT approaches are inadequate for addressing deep-seated knowledge biases and perceptual errors, we provide a comprehensive analysis of failure cases observed in the baseline Qwen-Omni model for generating CoT reasoning. 

\subsubsection{Dataset Composition Analysis}
We initiated our investigation with a systematic examination of the dataset's compositional characteristics. Our analysis reveals a significant concentration of musical instruments in the dataset, with violin (215 samples), drum (148 samples), and piano (131 samples) being the most prevalent categories. This data skew is critical, as a disproportionate number of model failures occur within this domain.

\begin{figure}[!t]
    \centering
    \includegraphics[width=1\linewidth]{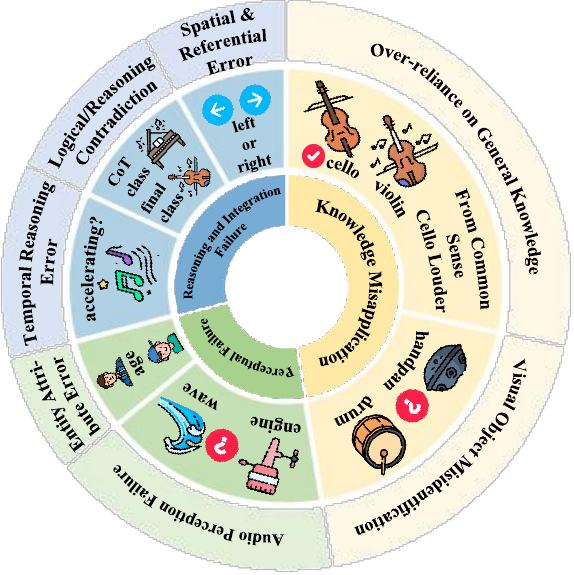}
    \caption{\textbf{A Taxonomy of Analyzed Failure Cases from the Baseline Qwen-Omni Model.} It illustrates the categorical distribution of errors within our curated analysis set. The significant proportion of Knowledge Misapplication and Perceptual Failure provides empirical evidence that standard SFT with CoT training is insufficient to address these deep-seated and foundational error types.
    }
    \label{reflective}
    \vspace{-0.5cm}
\end{figure} 

\subsubsection{Failure Taxonomy}
The primary focus of our investigation centers on developing a systematic taxonomy of model failures. Through rigorous analysis, we categorize the observed errors into three fundamental types: Knowledge Misapplication, Perceptual Failure, and Reasoning and Integration Failure, which as shown in Figure~\ref{reflective}.

\begin{figure*}[!t]
    \centering
    \includegraphics[width=1\linewidth]{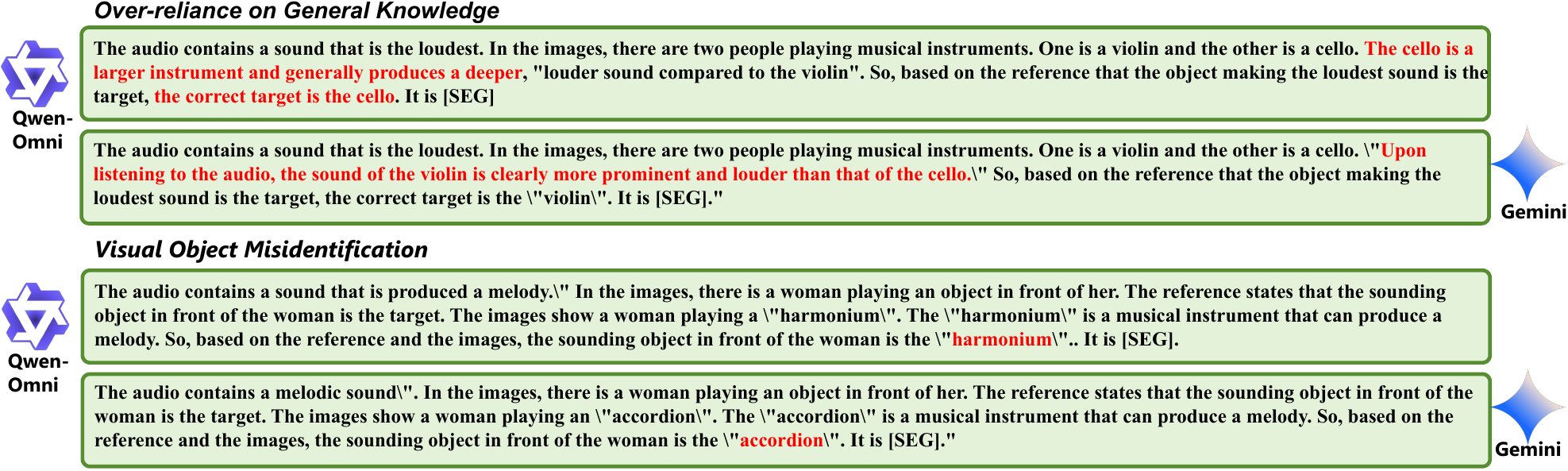}
    \caption{\textbf{Example of Knowledge Misapplication, which consists of Over-reliance on General Knowledge and Visual Object Misidentification.}
    }
    \label{Knowledge Misapplication}
\end{figure*} 

\begin{figure*}[!t]
    \centering
    \includegraphics[width=1\linewidth]{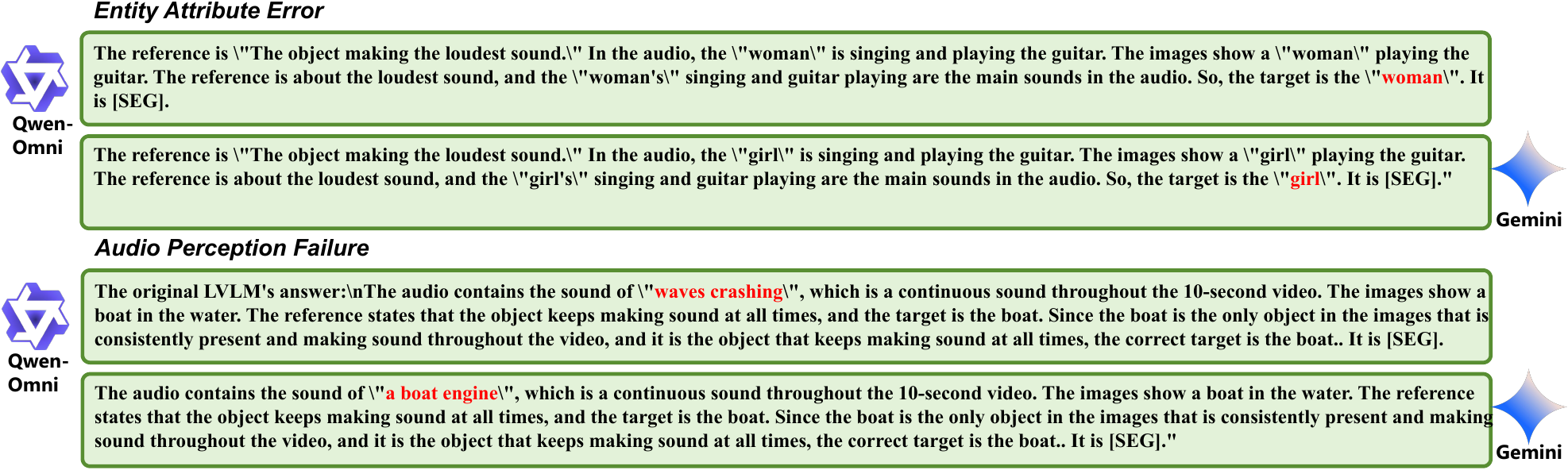}
    \caption{\textbf{Example of Perceptual Failure, which consists of Audio Perception Failure and Entity Attribute Error.}
    }
    \label{Perceptual Failure}
\end{figure*} 

\begin{figure*}[!t]
    \centering
    \includegraphics[width=1\linewidth]{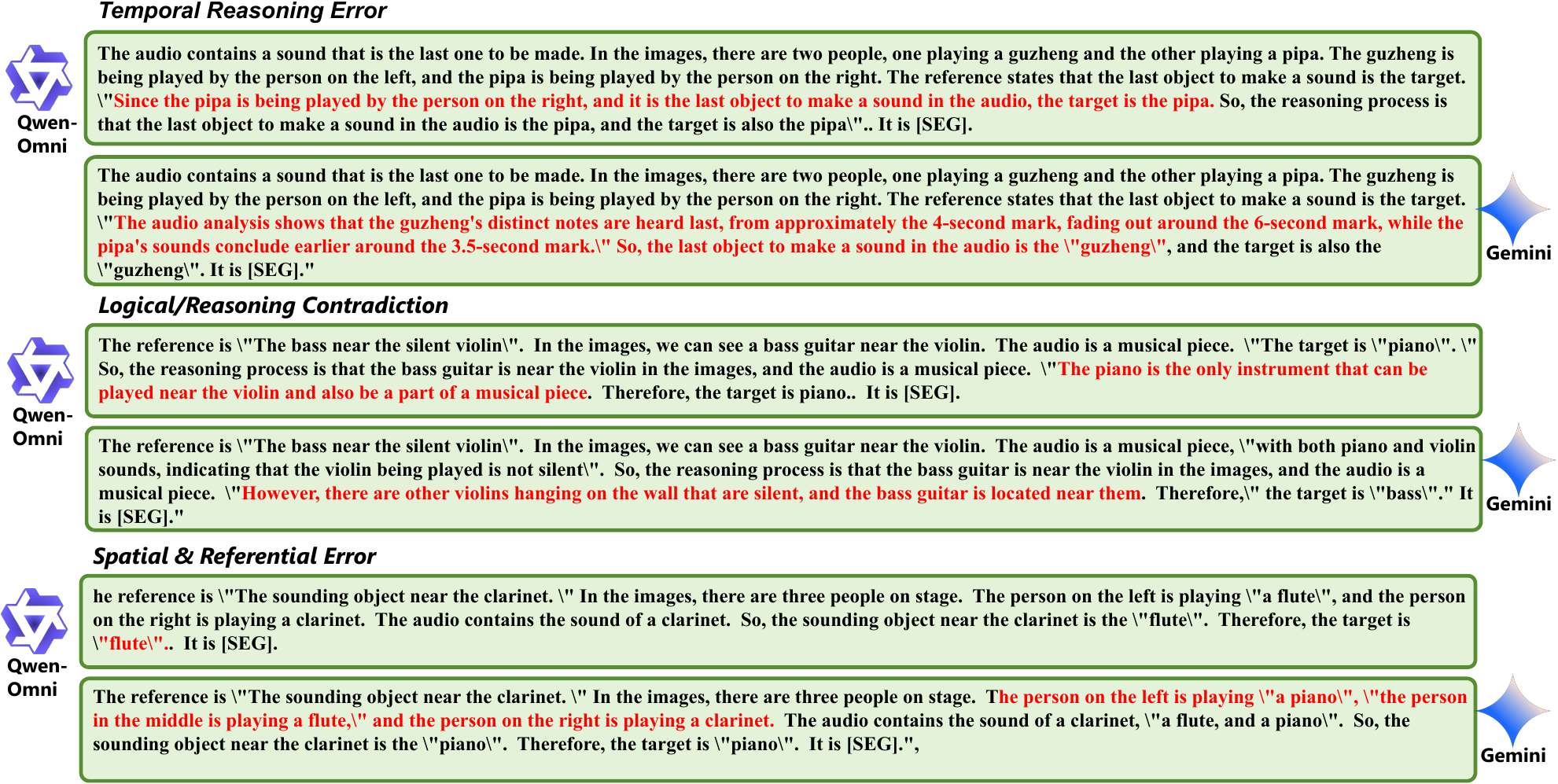}
    \caption{\textbf{Example of Reasoning and Integration Failure, which consists of Temporal Reasoning Error, Logical/Reasoning Contradiction and Spatial \& Referential Error.}
    }
    \label{Reasoning and Integration Failure}
\end{figure*} 

\paragraph{\textbullet\ Knowledge Misapplication}

Among these, our analysis revealed that Knowledge Misapplication is the most significant and frequent failure mode, thus becoming the central motivation (modify prior knowledge) for the corrective reflective-style training scheme proposed in our work. This primary issue is manifested in two distinct forms. Examples of Over-reliance on General Knowledge and Visual Object Misidentification are shown in Figure~\ref{Knowledge Misapplication}.

\textbf{Over-reliance on General Knowledge} is the most prevalent form, occurring in 428 samples, where the model use its ingrained ``common sense" priors instead of analyzing direct evidence from the input. A canonical example is its handling of queries about the relative loudness of a cello and a violin. In a vast majority of cases, the model consistently selects the cello as louder, providing the rationale that ``the cello is a larger instrument and usually has a louder sound." This general heuristic is misapplied in a context where the model should be analyzing the acoustic properties of the specific audio clip, in which the violin, as the lead melodic instrument, is often more prominent. 

\textbf{Visual Object Misidentification} (285 samples) is the second form of this error, wherein the model defaults to correctly recognizing specific objects, frequently confusing visually similar instruments such as a guitar with a ukulele or a handpan with a drum.

Taken together, these findings highlight a fundamental weakness: the model's inability to dynamically override its static prior knowledge with new, contextual evidence. This empirical justification underscores the necessity for the corrective reflective-style training, designed specifically to patch these flawed priors. 

\paragraph{\textbullet\ Perceptual Failure}

Perceptual failures can be broadly categorized into Audio Perception Failures and Entity Attribute Errors, which constitute the core issues we aim to address in Stage 2 by calibrating the model's fundamental perceptual capabilities. Examples of Audio Perception Failure and Entity Attribute Error are shown in Figure~\ref{Perceptual Failure}.

\textbf{Audio Perception Failure} refers to cases where the model directly misinterprets the audio content, with 251 such examples identified in our analysis. This is not a failure of common sense reasoning, but rather a pure perceptual error. For example, the model may confuse a female voice with a male voice, or misclassify an engine sound as ocean waves. These errors motivated us to introduce a ``reflective" reasoning mechanism, which prompts the model to compensate for perceptual inaccuracies.

\textbf{Entity Attribute Error} (103 samples) stems from visual perception, where the model struggles to correctly identify attributes such as age. For example, it may confuse a woman with a girl or a man with a boy. Although such misclassifications may not always directly affect reasoning outcomes, they can lead to ambiguity or confusion when multiple individuals are present in a scene.

\paragraph{\textbullet\ Reasoning and Integration Failure}

The third category of errors is Reasoning and Integration Failure, which primarily reflects deficiencies in the model's reasoning capabilities. Unlike perceptual errors and prior knowledge errors addressed in the ``reflective” training stage, these issues are mainly targeted in the GRPO stage. This category includes three major types: Temporal Reasoning Errors, Logical/Reasoning Contradictions, and Spatial \& Referential Errors. Example of these three major types is shown in Figure~\ref{Reasoning and Integration Failure}.

\textbf{Temporal Reasoning Error} (166 samples) refers to the model's failure in understanding the temporal sequence or rhythmic progression of auditory events. For example, it may incorrectly judge whether a sound is accelerating or decelerating, or whether the volume is increasing or decreasing. These errors often arise because the model fails to analyze the overall temporal trend across the entire audio clip.

\textbf{Logical/Reasoning Contradiction} (148 samples) occurs when the model's reasoning chain contains internal inconsistencies, or when the final answer contradicts intermediate reasoning steps. This may involve hallucinations, cases where the model identifies correct information but still provides an incorrect answer, or neglects crucial keywords present in the reference.

\textbf{Spatial \& Referential Error} (124 samples) refers to failures in comprehending spatial relationships, such as "left," "right," or "behind", and in correctly referring to an object among multiple candidates. The model may misidentify which object is being referred to, especially in scenes with similar entities or complex spatial layouts.

\begin{figure}[!t]
    \centering
    \includegraphics[width=1\linewidth]{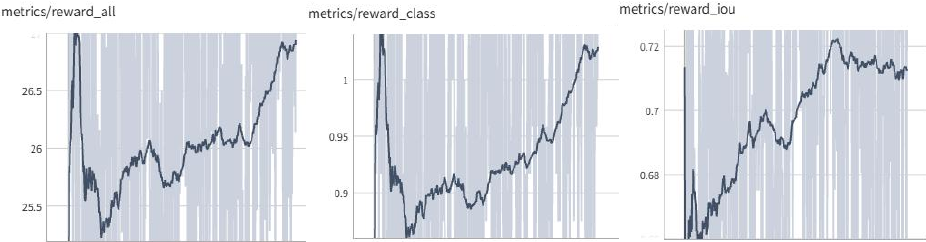}
    \caption{\textbf{ The training curves for the reward function in GRPO.} The subplots correspond to the overall reward (metrics/reward\_all), class reward (metrics/reward\_class), and IoU reward (metrics/reward\_iou), respectively. The x-axis represents training steps.}
    \label{reward_curve}
\end{figure} 

\begin{figure}[!t]
    \centering
    \includegraphics[width=1\linewidth]{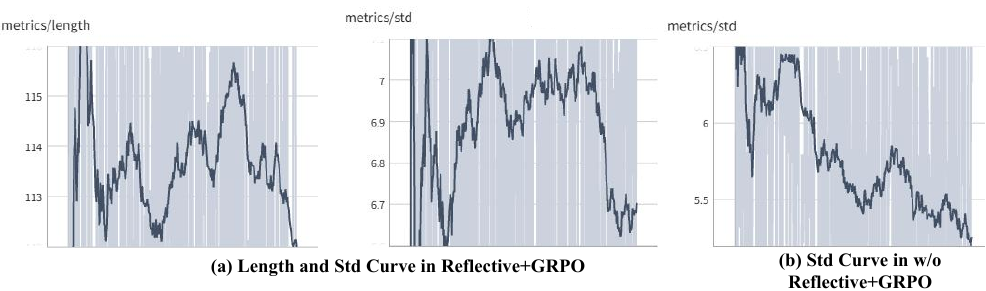}
    \caption{\textbf{The training curves for the reasoning length, and intra-group std in GRPO with and w/o reflective-style training. } The x-axis represents training steps.}
    \label{length_curve}
\end{figure} 

\subsection{Training curve of GRPO}
To evaluate the training process of GRPO stage, we present the training curves for the reward function, reasoning length, and intra-group standard deviation (std) as shown in Figure~\ref{reward_curve}. During training, the IoU reward, class reward, and final reward all exhibit an increasing trend, which validates the effectiveness of our GRPO training method. Furthermore, the reasoning length and intra-group std do not decline rapidly in the early stages, indicating that the model explores effective reasoning paths. To further demonstrate the effectiveness of ``reflective" training, we compare the full model with a variant trained without Stage 2  as shown in Figure~\ref{length_curve} . In the absence of corrective reflective-style training, the std curve declines rapidly at an early stage (Figure~\ref{length_curve} (b)). This result demonstrates that introducing the ``reflective" path provides the model with more space to explore—namely by increasing the policy entropy \cite{wang2025beyond, cui2025entropy}—thus leading to better performance.

\section{More Experiments}
\subsection{SFT Segmentation Token}
For MLLM-based segmentation approaches, existing methods primarily adopt two distinct formats: the \verb|[SEG]| token approach and the learnable token approach. While our main implementation utilizes the \verb|[SEG]| token format, we also conduct a comprehensive evaluation of the learnable token methodology to assess its effectiveness.
The learnable token approach represents a method that preserves the original vocabulary without modification, instead appending specialized tokens after the reasoning process to extract segmentation information. As demonstrated in Table~\ref{learnable}, we compare our approach against two variants employing learnable tokens: ``seg+MLLM" (segmentation training followed by reasoning training) and ``MLLM+seg" (reasoning training followed by segmentation training).
The experimental results reveal that our \verb|[SEG]| token approach without CoT formatting significantly outperforms both learnable token variants by a substantial margin, demonstrating the superiority of our proposed methodology.

\begin{table}[h]
    \setlength{\belowcaptionskip}{0.2cm}   
    \renewcommand{\arraystretch}{1}      % 稍微减少行间距，之前1.5太大了
    \renewcommand{\tabcolsep}{5pt}         % 稍微减少列间距，10pt太宽
    \footnotesize                                    % 改为small，比footnotesize稍大一点
    \centering
	\begin{tabular}{l@{\hspace{1.5em}}ccc@{\hspace{1em}}ccc}
		\toprule
		\multirow{2}{*}{\textbf{Method}} & \multicolumn{3}{c}{\textbf{Seen}}  & \multicolumn{3}{c}{\textbf{Unseen}} \\
		\cmidrule(lr){2-4} \cmidrule(lr){5-7} 
		& \textbf{$\mathcal{J}$} & \textbf{$\mathcal{F}$} & \textbf{$\mathcal{J}\!\&\!\mathcal{F}$} & \textbf{$\mathcal{J}$} & \textbf{$\mathcal{F}$} & \textbf{$\mathcal{J}\!\&\!\mathcal{F}$} \\
        \midrule
        seg+MLLM &46.5 &55.9  &51.2 &52.3 &59.4  &55.9 \\
        MLLM+seg &15.0 &19.1 &17.1  &33.9 &41.9 & 37.9\\
        % SFT+w/o MLLM &39.2 &48.7 &44.0 &34.0  &46.3 &40.2\\
        \rowcolor{gray!20} SFT+w/o CoT &\textbf{54.1} &\textbf{64.0} &\textbf{59.1} &\textbf{64.2} &\textbf{71.9} &\textbf{68.1}\\
		\bottomrule
	\end{tabular}
    \vspace{-3mm}
    \caption{Ablation study of learnable token and [SEG] token in the SFT training.} 
    \label{learnable}
    \vspace{-4mm}
\end{table}

\subsection{CoT Format Analysis}
We further examine the impact of CoT format variations on model performance. In our main implementation, we introduce a reasoning process that analyzes audio, visual, and reference information before generating the final answer through concise reasoning steps, which we term ``flexible CoT." This approach allows for adaptive sequencing and flexible analytical language. To evaluate this design choice, we compare it against a ``fixed CoT" format that employs a rigid structure with standardized analytical language.

The comparison between flexible and fixed CoT formats is presented below and we also compare their performance in SFT (Table~\ref{cot}). We hypothesize that the fixed format introduces unnecessary learning complexity in SFT, and the complex reasoning chains may cause the model to overfit on linguistic reasoning patterns, thereby reallocating its representational capacity away from fine-grained perception and segmentation. Furthermore, rigid formatting constraints may limit the model's exploratory capacity during the GRPO optimization stage.

\begin{tcolorbox}[
enhanced,
colback=gray!8,
colframe=black!60,
boxrule=1pt,
left=6pt,
right=6pt,
top=4pt,
bottom=4pt
]
\textit{
\textbf{Flexible Format}: ``The audio contains a sound that is produced by an object being played by the man. In the images, we can see a man playing a cello and a woman playing a violin. The reference states that the object making the sound is played by the man. Since the cello is the instrument being played by the man in the images, and it's the object that makes a sound in the audio, the target is cello." \\
\textbf{Fixed Format}: ``Reasoning: 1) Analyze the Reference: The reference statement is "The object making a sound by being played by the man." This directs the focus to identifying the object that the man is playing and making a sound with. 2) Analyze the Audio: The audio clip features two main sounds: a woman playing the violin and a man playing the cello. 3) Analyze the Images: The images show a woman playing the violin and a man playing the cello. The man is clearly playing the cello, as he is holding and using the cello in a way that is consistent with playing it. 4) Connect Audio and Visuals: The man is the one playing the cello in the images, and the cello is the object that is making a sound in the audio. 5) Deduce the Target: Since the reference asks for the object that the man is playing and making a sound with, and the audio and images both show the man playing the cello, the correct target is the cello." \\}
\end{tcolorbox}

\begin{table}[h]
    \setlength{\belowcaptionskip}{0.2cm}   
    \renewcommand{\arraystretch}{1}      
    \renewcommand{\tabcolsep}{5pt}         
    \footnotesize                            
    \centering
	\begin{tabular}{l@{\hspace{1.5em}}ccc@{\hspace{1em}}ccc}
		\toprule
		\multirow{2}{*}{\textbf{Method}} & \multicolumn{3}{c}{\textbf{Seen}}  & \multicolumn{3}{c}{\textbf{Unseen}} \\
		\cmidrule(lr){2-4} \cmidrule(lr){5-7} 
		& \textbf{$\mathcal{J}$} & \textbf{$\mathcal{F}$} & \textbf{$\mathcal{J}\!\&\!\mathcal{F}$} & \textbf{$\mathcal{J}$} & \textbf{$\mathcal{F}$} & \textbf{$\mathcal{J}\!\&\!\mathcal{F}$} \\
        \midrule
        Fixed Format &61.0 &70.1 &65.6 &66.4 &72.7 &69.6 \\
        \rowcolor{gray!20} Flexible Format &\textbf{61.4} &\textbf{70.6} &\textbf{66.0} &\textbf{67.1} &\textbf{73.4} &\textbf{70.3}\\
		\bottomrule
	\end{tabular}
    \vspace{-3mm}
    \caption{Ablation study of fixed and flexible format for CoT.} 
    \label{cot}
    \vspace{-4mm}
\end{table}

\begin{figure*}[!t]
    \centering
    \includegraphics[width=1\linewidth]{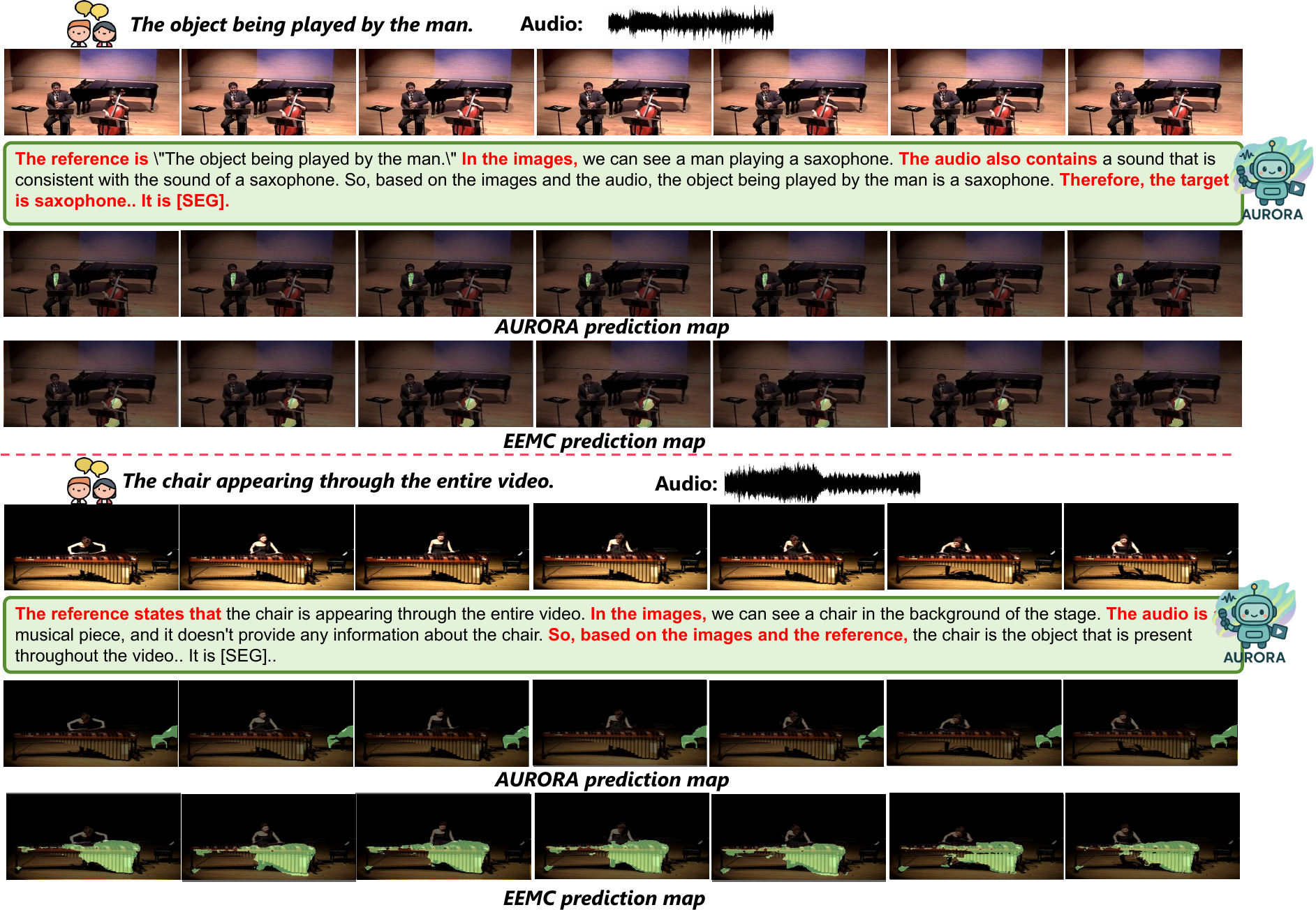}
    \vspace{-0.7cm}
    \caption{\textbf{The visualization results of the referred objects for the image-based examples in the Ref-AVS.} }
    \label{image-based}
    \vspace{-0.5cm}
\end{figure*} 

\section{More Comparison Show}
\begin{figure*}[!t]
    \centering
    \includegraphics[width=1\linewidth]{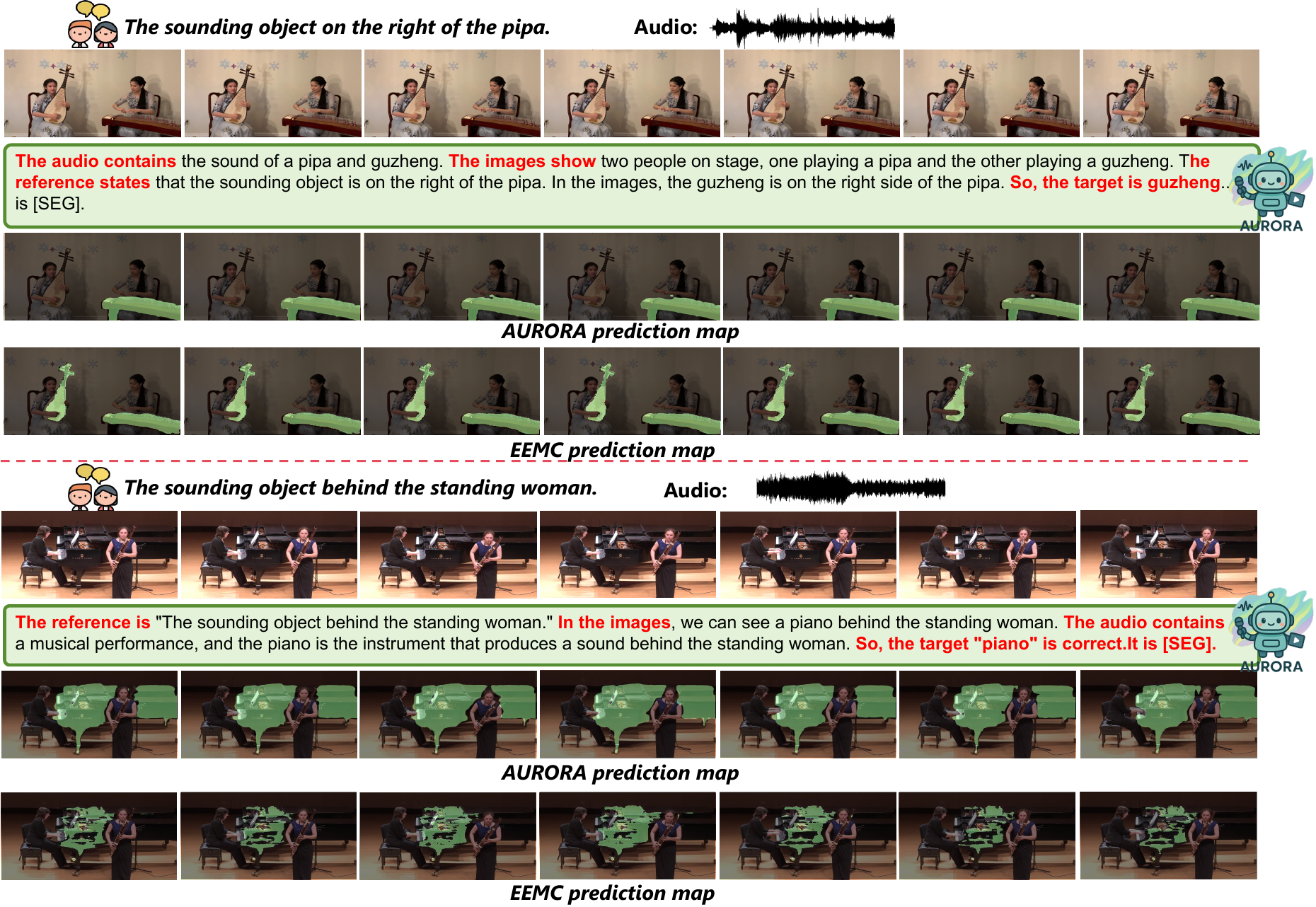}
    \vspace{-0.7cm}
    \caption{\textbf{The visualization results of the referred objects for the spatial-based examples in the Ref-AVS.} }
    \label{spatial}
    \vspace{-0.5cm}
\end{figure*}

\begin{figure*}[!t]
    \centering
    \includegraphics[width=1\linewidth]{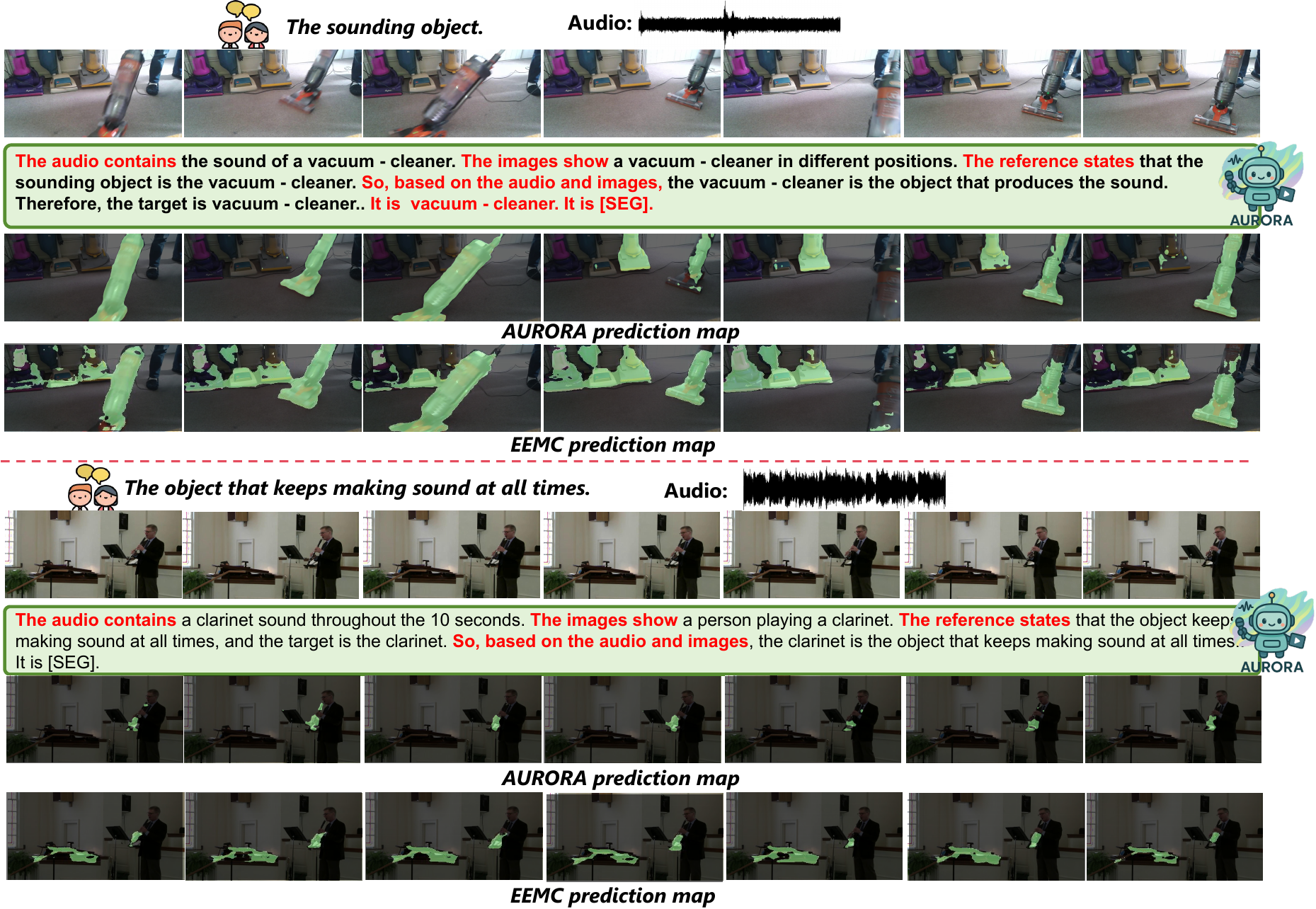}
    \vspace{-0.7cm}
    \caption{\textbf{The visualization results of the referred objects for the audio-based examples in the Ref-AVS.} }
    \label{audio}
    \vspace{-0.5cm}
\end{figure*}

To ensure a fair comparison, our evaluation is focused on methods with publicly available pre-trained weights. We further present more visual comparison against EEMC \cite{wang2024ref} on the Ref-AVS benchmark dataset \cite{wang2024ref} to highlight our model's effectiveness. As shown in Figures~\ref{image-based}, Figures~\ref{spatial}, and Figures~\ref{audio}, our model demonstrates exceptional performance in various challenging scenarios—including image, spatial, and audio-based cases—where existing methods often struggle.

\end{document}